\documentclass{article}
\usepackage{spconf,amsmath,graphicx}

\usepackage{enumitem}
\setlist{nosep, leftmargin=14pt}

\usepackage{mwe} % to get dummy images
\usepackage{cite}
\usepackage{amsmath}
\usepackage{amssymb}
\usepackage{algorithmic}
\usepackage{algorithm}
\usepackage[thinc]{esdiff}
\usepackage{svg}
\usepackage{wrapfig}
\usepackage{graphicx}
\usepackage{caption}
\usepackage{subcaption}
\usepackage{array}
\usepackage{multirow}

\DeclareMathOperator{\encoder}{\psi}
\DeclareMathOperator{\decoder}{\phi}

\newcommand\blfootnote[1]{%
  \begingroup
  \renewcommand\thefootnote{}\footnote{#1}%
  \addtocounter{footnote}{-1}%
  \endgroup
}

\title{Measuring Feature Dependency of Neural Networks by Collapsing Feature Dimensions in the Data Manifold}
\name{Yinzhu Jin$^{\star}$\qquad Matthew B. Dwyer$^{\star}$\qquad P. Thomas Fletcher$^{\dagger,\star}$}
\address{$^{\star}$ Department of Computer Science, University of Virginia, Charlottesville, VA, USA\\
     $^{\dagger}$ Department of Electrical and Computer Engineering, University of Virginia,\\ Charlottesville, VA, USA}

\begin{document}
%\ninept
%
\maketitle
\begin{abstract}
This paper introduces a new technique to measure the feature dependency of neural network models.
The motivation is to better understand a model by querying whether it is using information from human-understandable features, e.g., anatomical shape, volume, or image texture. Our method is based on the principle that if a model is dependent on a feature, then removal of that feature should significantly harm its performance.
A targeted feature is ``removed'' by collapsing the dimension in the data distribution that corresponds to that feature. We perform this by moving data points along the feature dimension to a baseline feature value while staying on the data manifold, as estimated by a deep generative model.
Then we observe how the model's performance changes on the modified test data set, with the target feature dimension removed.
We test our method on deep neural network models trained on synthetic image data with known ground truth, an Alzheimer's disease prediction task using MRI and hippocampus segmentations from the OASIS-3 dataset~\cite{lamontagne2019oasis}, and a cell nuclei classification task using the Lizard dataset~\cite{graham2021lizard}.
\end{abstract}
%
%\begin{keywords}
%One, two, three, four, five
%\end{keywords}

\blfootnote{

Copyright \copyright 2024 IEEE

Personal use of this material is permitted. Permission from
IEEE must be obtained for all other uses, in any current or future media, including reprinting/republishing this material for
advertising or promotional purposes, creating new collective
works, for resale or redistribution to servers or lists, or reuse
of any copyrighted component of this work in other works.

Published in: 2024 IEEE 21st International Symposium on Biomedical Imaging (ISBI)}

\section{Introduction}

Deep neural networks (DNNs) have shown great success in many medical imaging tasks but lack transparency in their decision-making processes. This is particularly problematic in medical applications of deep learning for at least two reasons. First, for a deep learning system to be trustworthy when making health care decisions, it is critical that its decision rules be explainable and plausible to a medical expert. Second, in clinical research it is often important to derive understanding of a biological process. Both scenarios benefit from the ability to explain the features a DNN is using in terms that a human expert can understand. In this work, we are specifically interested in evaluating whether certain target features are indeed used by an existing DNN. This is in contrast to explainability methods that strive to extract features being computed by a DNN and present them in a hopefully interpretable fashion~\cite{sundararajan2017axiomatic, hesse2021fast, selvaraju2017grad, bach2015pixel,goyal2019counterfactual, kanamori2020dace}. Rather, we wish to take features that are known to be interpretable---and important in a particular domain---and query if those features are being used by the DNN. For example, it is a well-established fact that Alzheimer's disease (AD) causes atrophy of the hippocampus. If we are given a DNN that classifies AD patients versus healthy subjects from magnetic resonance imaging (MRI), we would want to know if that classifier is in some way using the volume of a subject's hippocampi in its decision rule.

To address this problem, one prevailing option is to locally approximate the nonlinear decision boundary with an easier to interpret linear approximation, either in data space~\cite{ribeiro2016should, ahern2019normlime} or latent space~\cite{kim2018interpretability}. Yet, most interpretable features of interest are nonlinear functions, which can be lost by linear approximations. Jin et al.~\cite{jin2022feature} tackles non-linearity by interpreting neural networks in terms of the alignment between gradients of a classifier and gradients of a feature along the gradient flow of classifiers. However, the magnitude of the gradient alignment can itself be hard to connect to how important that feature is to the classifier. Closer in nature to our proposed approach, CaCE~\cite{goyal2019explaining} explores the impact of modifying discrete concepts of the input by utilizing conditional generators. We extend it to continuous features and draw comparison with our proposed method in the experiments.

Our proposed method aims to ``remove'' a feature from a test dataset and measure how much this negatively affects the test performance of the DNN. When the features are merely subsets of the original input dimensions, they can be easily masked~\cite{chen2023algorithms}. However, this is often not useful for imaging data, where the original input dimensions correspond to individual pixel/voxel values. When the target feature is a complex, nonlinear function of the input image, removing it from the data, while keeping it valid and realistic, is not trivial. We propose to do this by modeling {\em feature collapse} as an integral curve of the target feature's gradient vector field in the latent space of a generative model that has learned the data distribution. By restricting to the data manifold estimated by the generative model, we ensure that other characterizing features of the data are preserved. We can eliminate a target feature from the test data by manipulating each data point to have a common constant value for this feature (e.g., adjust MR images so that they all have equal hippocampal volume).

\begin{figure}[t]
    \centering
    \includegraphics[width=0.4\textwidth]{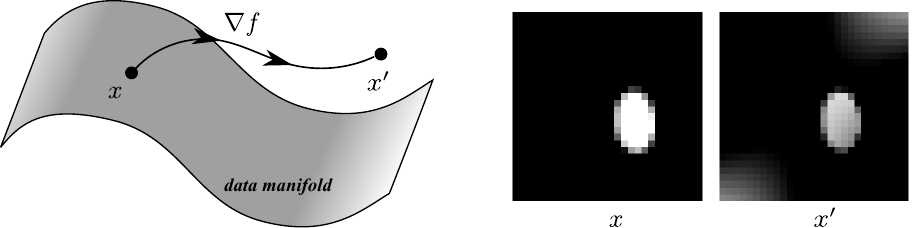}
    \caption{Illustration of integrating a feature gradient in the ambient data space, where the resulting endpoint, $x'$ lands off of the data manifold (left). Example of this effect using aspect ratio of an ellipse as the feature (right).}
    \label{fig:flow}
%    \vspace{-3mm}
\end{figure}

\section{Methods}
We consider a neural network classifier as a mapping $g: \mathbb{R}^D \to \mathbb{R}^K$, where an input image is considered as a point $x \in \mathbb{R}^D$, and the corresponding output is the vector of assigned log class probabilities, $\ln p(y = k \mid x), k = 1, \ldots, K$. We define a feature as a differentiable function $f: \mathbb{R}^D \to \mathbb{R}$. To measure the dependency of the classifier $g$ on the feature $f$, we propose to observe the change in performance of $g$, e.g., accuracy, when the feature $f$ is ``removed''. Given the original test dataset $X$, we modify each point in $X$ by collapsing the dimension corresponding to the feature $f$. The modified test set with the feature $f$ collapsed is denoted $X_{\overline{f}}$. We then test the classifier $g$ on this new test dataset and compare the performance on $X_{\overline{f}}$ to the original performance on $X$. The dependency of $g$ on the feature $f$ is reflected by how much the performance drops.

Collapsing a feature dimension is generally a non-trivial task. If the feature is a linear function of the input, i.e., $f(x) = v \cdot x$, where $v \in \mathbb{R}^D$ is a constant unit vector, then collapsing the feature dimension is simply projection onto the orthogonal complement of $v$. In other words, the collapsing operation in the linear case is given by
$x_{\overline{f}} = x - (x \cdot v) v.$ The resulting collapsed data points will have constant feature value, $f(x_{\overline{f}}) = 0$, so the information from $f$ is effectively removed. The more general case, where $f$ is a nonlinear function, is more complicated. We might consider moving data points along integral curves of the gradient of $f$ until we arrive at a constant value for $f$. That is, we integrate the ordinary differential equation
\begin{equation}
\label{eq:integrate}
\frac{dc}{dt}(t) = \nabla{f}(c(t)),
\end{equation}
with initial conditions $c(0) = x$, and stopping when $f(c(t)) = b$ for some predetermined baseline value $b$. Note that this is analogous to the linear case, where orthogonal projection moves along the constant gradient field, $\nabla f = v$. Also note that the integration of \eqref{eq:integrate} may need to be forward or backward in $t$, depending on whether the inital feature value, $f(x)$, is above or below the baseline $b$.

There is a serious drawback to this strategy of moving a data point along the integral curves of $\nabla f$, which is that the integral curve may move outside of the data distribution. More specifically, if we think of our data distribution as lying on a lower-dimensional manifold in the data space, the integral curves of $\nabla f$ may leave the data manifold. This is illustrated in Fig.~\ref{fig:flow}. Thus, moving along the integral curves of $\nabla f$ may produce invalid data, i.e., data that does not look like realistic samples from the data distribution. As an example, imagine we have a dataset of images of white ellipses on a black background, and we can compute the aspect ratio of the ellipse as our feature $f$. As shown on the right side of Fig.~\ref{fig:flow}, if we try to change the aspect ratio of an ellipse image directly by moving along the gradient direction of this feature in the ambient image space, then we produce an image that does not look like an ellipse with adjusted aspect ratio. This is because we have moved off of the manifold of valid ellipse images.

\begin{figure}[t]
    \centering
    \includegraphics[width=0.4\textwidth]{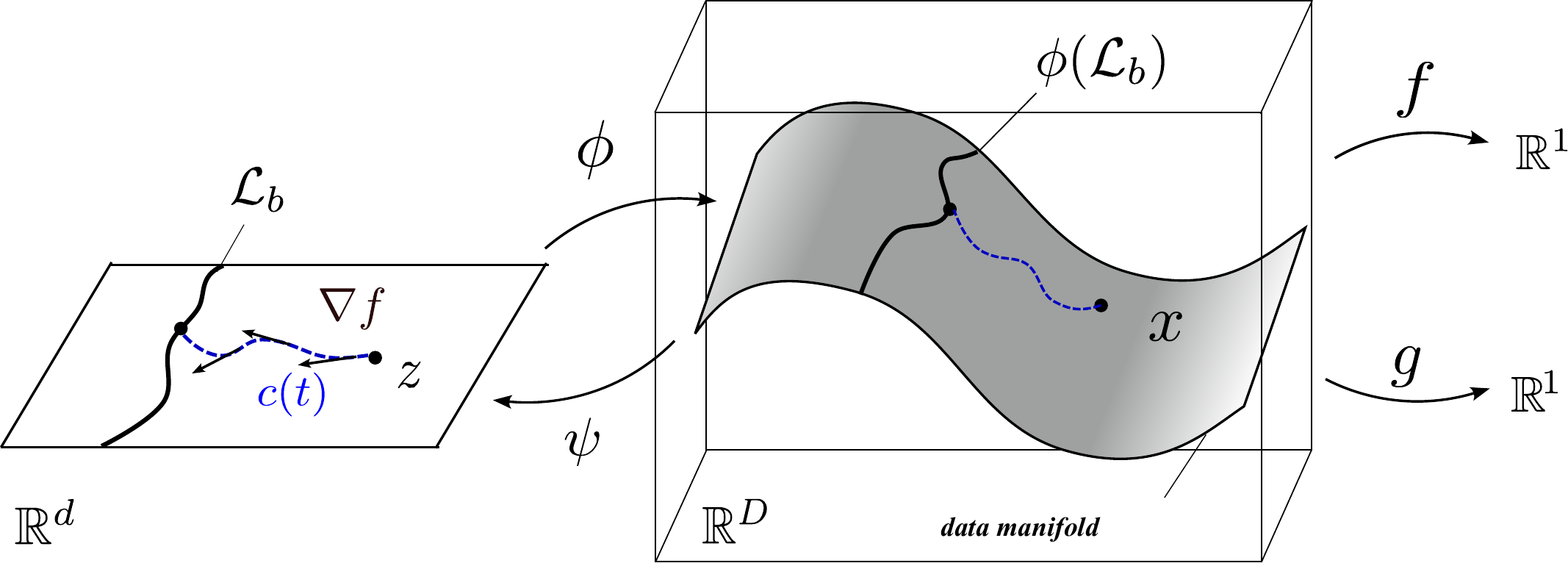}
    \caption{Illustration of the proposed feature collapsing method.}
    \label{fig:overall}
%    \vspace{-3mm}
\end{figure}

To handle this, we propose to first train a deep generative model to learn the data distribution and then restrict movements along our feature gradient to remain on the estimated data manifold. In this work we use a variational autoencoder (VAE)~\cite{kingma2013auto} with encoder $\psi : \mathbb{R}^D \rightarrow \mathbb{R}^d$ and decoder $\phi : \mathbb{R}^d \rightarrow \mathbb{R}^D$, where $\mathbb{R}^d$ ($d < D$) is the latent space. Given a data point $x \in \mathbb{R}^D$, we first encode it to produce a latent representation, $z = \encoder(x)$. Now, we proceed with the same strategy to collapse the feature $f$ by moving along the gradient direction, but now we constrain this to be the gradient of $f$ restricted to the estimated data manifold. The feature $f$ restricted to the VAE manifold is given by $f \circ \decoder$, and the integral curve of the gradient is now
\begin{equation}
\label{eq:integral_latent}
\frac{dc}{dt}(t) = \nabla_z (f \circ \decoder (c(t))) = D\decoder(c(t))^T \nabla_x f(\decoder(c(t))),
\end{equation}
where $c(t)$ is a curve in the latent space, $\mathbb{R}^d$, $D\decoder$ is the Jacobian matrix for 
$\decoder$, and $\nabla_z$, $\nabla_x$ are gradients with respect to a latent $z$ or data $x$, respectively. Note that the multiplication by $D\decoder^T$ in \eqref{eq:integral_latent} comes from the chain rule and is computed as a backpropagation through the decoder, 
$\decoder$.

We start integrating \eqref{eq:integral_latent} at the encoded input data point, $c(0) = z$, and we integrate (forward or backward) until we reach a desired baseline feature value at some time $T$. The end result is a point $z_{\overline{f}} = c(T)$ that lies on the baseline level set for $f$, $\mathcal{L}_{b} = \{z \mid f(\phi(z))=b\}$. This corresponds to a data point with the $f$ feature collapsed, that is, it produces $x_{\overline{f}} = \decoder(z_{\overline{f}})$, such that $f(x_{\overline{f}}) = b$. The overall process is illustrated in Fig.~\ref{fig:overall}. In practice, we perform the gradient vector field integration with a discrete Euler integration step, which is summarized in Algorithm~\ref{alg:collapse}.

%\begin{wrapfigure}[13]{r}{0.5\textwidth}
%\vspace{-14mm}
%\begin{minipage}{0.45\textwidth}
\begin{algorithm}[t]
\caption{Feature Collapse}
\label{alg:collapse}
\begin{algorithmic}
\REQUIRE A data point $x$
\ENSURE Data point $x_{{\overline{f}}}$ returned
\STATE $z \gets \encoder(x)$ and $x' \gets \decoder(z)$
\STATE $s \gets \mathrm{sign}(f(x') - b)$ 
\WHILE{$s \cdot (f(x')-b) > 0$}
    \STATE $v \gets D\decoder(z)^T \nabla f(x')$
    \STATE $z \gets z - s \alpha \cdot v$
    \STATE $x' \gets \decoder(z)$
\ENDWHILE
\RETURN $x'$
\end{algorithmic}
\end{algorithm}

\begin{figure}[t]
\centering
    \begin{subfigure}[b]{0.49\textwidth}
         \centering
         \includegraphics[width=\textwidth]{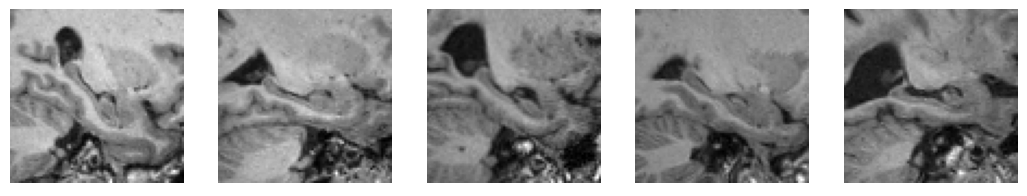}
    \end{subfigure}
    \begin{subfigure}[b]{0.49\textwidth}
         \centering
         \includegraphics[width=\textwidth]{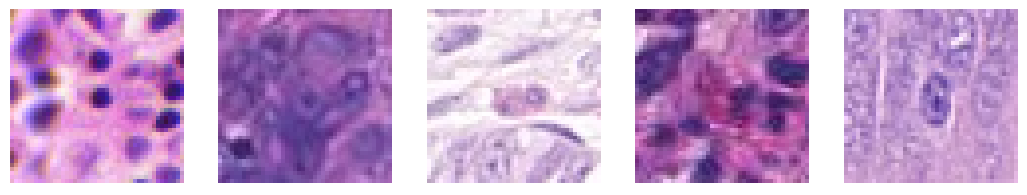}
    \end{subfigure}
    \caption{Sample cropped images from OASIS-3 (top) and Lizard (bottom).}
%    \vspace{-3mm}
\end{figure}

\section{Experiments}
\subsection{Setup}
We test our method on three image datasets: a synthetic dataset of binary ellipse images, hippocampi from MRI in OASIS-3~\cite{lamontagne2019oasis}, and cell nuclei histology images from the Lizard dataset~\cite{graham2021lizard}.
All experiments are implemented using PyTorch~\cite{paszke2017automatic}.
For each classification task, we perform $5$-fold cross validation. We evaluate the dependency of each classifier on certain interpretable features using our proposed feature collapse method (Algorithm~\ref{alg:collapse}) applied to the test set, and compare the results to CaCE scores~\cite{goyal2019explaining}.

The ellipse dataset contains 10,000 grayscale images of white ellipses on black backgrounds with five varying generative factors: $x$ and $y$ position, size, rotation angle, and aspect ratio. We assign them to two classes separated by only aspect ratio.
For the hippocampi dataset, we cropped a region of interest around the left and right hippocampi in T1w MRI from 925 different subjects in the OASIS-3 dataset~\cite{lamontagne2019oasis}, masked out the background voxels using Freesurfer segmentations~\cite{fischl2002whole} and applied Gaussian blur. The task is to classify AD versus healthy subjects.
The cell nuclei histology data is derived from the Lizard dataset~\cite{graham2021lizard}. We cropped the images around each nucleus, 2000 from each of the six annotated nuclei types, masked out other areas and applied Gaussian filter. 
For each dataset we trained a deep convolutional classifier for the task, a plain convolutional VAE to learn the data manifold for our method, and a conditional VAE to implement CaCE.

%\subsection{Interpretable Features}
We have chosen several interpretable features to test for classifier dependency as listed in Table \ref{results}. The aspect ratio is calculated by taking the ratio of major and minor eigenvalues of the second-order moments.
%We expect to see the model depends on the aspect ratio but not any other feature.
%For the hippocampus dataset, we compute the hippocampus volume, aspect ratio, and average voxel brightness.
%For the nucleus dataset, size, aspect ratio, saturation and hue are used. The saturation and hue are averaged over the foreground pixels.
For every dataset, we also included ten random linear features as baselines.
%, defined as $f_{v}(x) = x \cdot v$, where $v$ is a random $D$-dimensional unit vector, uniformly sampled on the unit sphere. 
Since CaCE was originally proposed for discrete concepts, we extend it by using a low and a high feature value to represent each feature and generate two sets of random samples using the conditional VAE. Different choices of feature values have been experimented as shown in the Table \ref{results}.

\begin{figure}
\centering
    \begin{subfigure}[b]{0.43\textwidth}
         \centering
         \includegraphics[width=\textwidth]{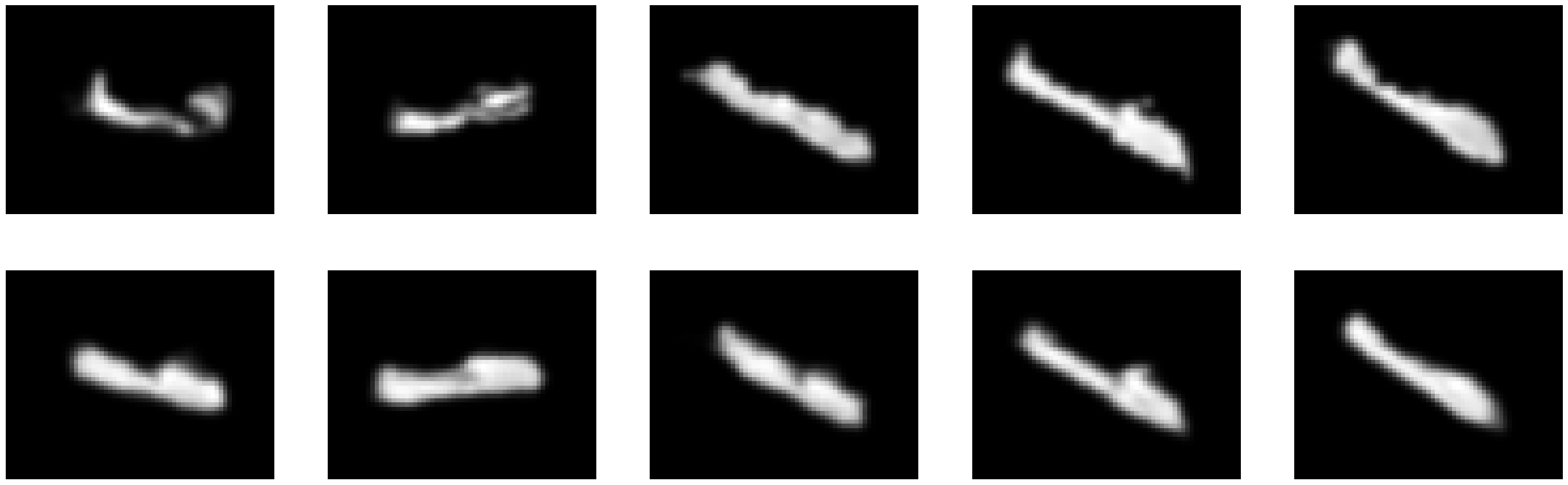}
         \caption{Volume in the hippocampus dataset}
    \end{subfigure}
    \begin{subfigure}[b]{0.43\textwidth}
         \centering
         \includegraphics[width=\textwidth]{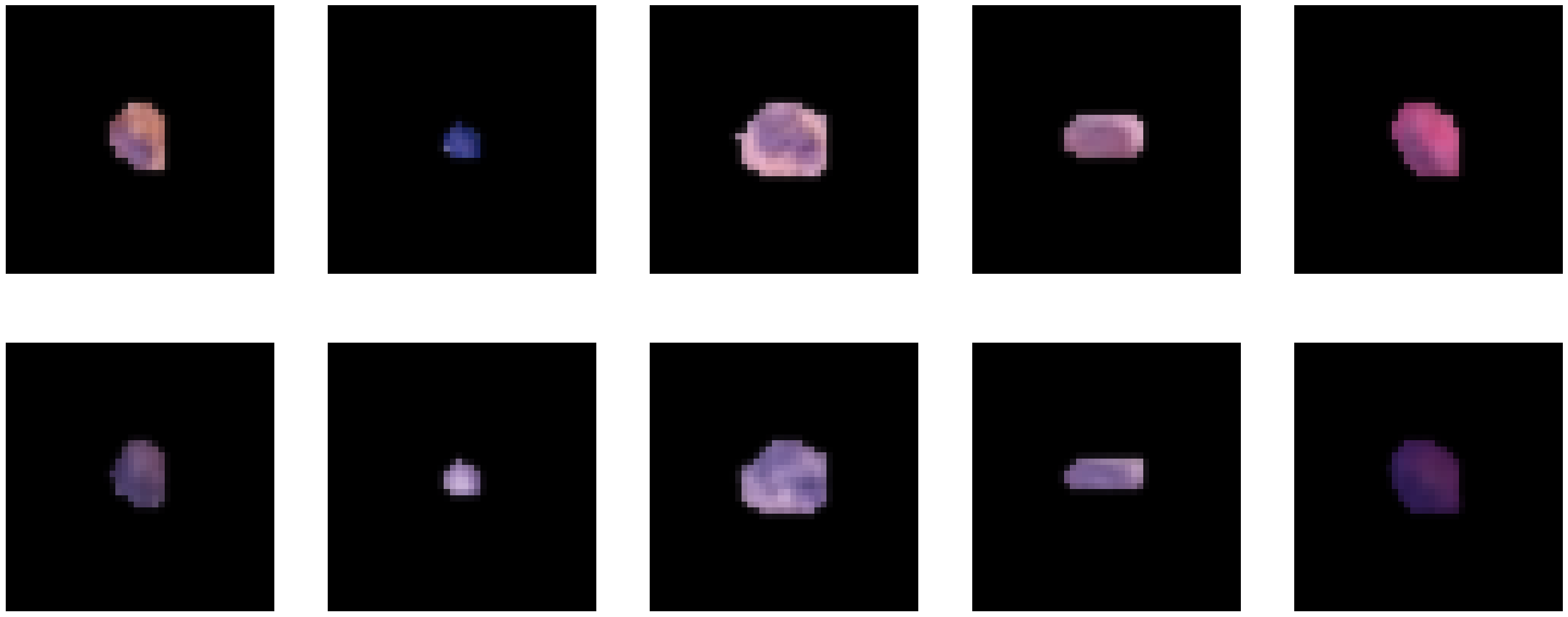}
         \caption{Hue in the cell nucleus dataset}
    \end{subfigure}
    \caption{Original samples (top rows) and samples after collapsing the given feature dimension (bottom rows).}
    \label{fig:feature}
%    \vspace{-4mm}
\end{figure}

\begin{figure}
    \centering
    \includegraphics[width=0.4\textwidth]{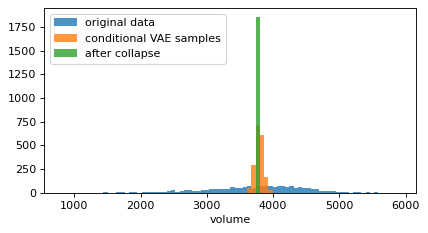}
    \caption{Hippocampi volume histogram of the original data, samples generated from conditional VAE and the samples after collapse along volume feature dimension}
    \label{fig:hist}
%    \vspace{-6mm}
\end{figure}
\subsection{Results}
First, we visualize the feature collapsing results in Fig.~\ref{fig:feature} for two examples: hippocampus volume and cell nucleus color hue. Every feature is collapsed to the mean value over the dataset. The altered images remain realistic, while the selected feature is successfully changed.
In Figure \ref{fig:hist}, we illustrate the distributions of volume feature values for three datasets: the original hippocampus data, samples generated by a conditional VAE with volume conditions set to the same baseline value, and samples generated by our method, which collapses the volume feature dimension. Our method demonstrates superior precision in constraining feature values around baseline values compared to the conditional VAE. Specifically, our generated samples consistently exhibit a feature value standard deviation less than $2\%$ of the original dataset.
In Table \ref{tab:accuracy}, we report the performance of the evaluated neural network models on the original dataset, which is averaged over 5 folds.
%For the random linear features, the mean value is calculated over all the ten sampled features.
The balanced accuracy is used for the hippocampus dataset. To verify that the reconstruction quality of the VAE is not affecting the classifier performance significantly, we also test the model on the reconstructed dataset, which shows a slight impact on the performance.

\begin{figure}[t]
\centering
\captionof{table}{Accuracies of evaluated classifiers on the original and reconstructed data. (mean $\pm$ std)}\label{tab:accuracy}
%\vspace{-4mm}
\begin{tabular}{|c|c|c|}
    \hline
     Dataset & original data & reconstructed data \\
     \hline
     Ellipse & 0.872 ($\pm$ 0.010) & 0.854 ($\pm$ 0.013) \\
     \hline
     Hippocampus & 0.821 ($\pm$ 0.020) & 0.816 ($\pm$ 0.028) \\
     \hline
     Nucleus & 0.606 ($\pm$ 0.004) &  0.581 ($\pm$ 0.002) \\
     \hline
\end{tabular}
%\vspace{-4mm}
\end{figure}

\begin{figure}[t]
\centering
\captionof{table}{The results of our methods: accuracy after collapse (AAC) along each feature dimension with features of substantial performance drops (relative to RLF) shown in bold; and the CaCE scores with two different sets of percentiles to represent low and high feature values.}\label{results}
%\vspace{-4mm}
\begin{subtable}[t]{0.485\textwidth}
\centering
\caption{Ellipse results}\label{ellipse}
\begin{tabular}{ | l | c | c |c|} 
  \hline
  \multirow{2}{1em}{Feature}& \multirow{2}{6em}{AAC (ours) (mean $\pm$ std)} &\multicolumn{2}{c|}{CaCE score} \\
  \cline{3-4}
  & & $25\%/75\%$ & $5\%/95\%$\\
  %\hline
  %Original &  0.872 ($\pm$ 0.010) \\
  %Reconstructed &  0.854 ($\pm$ 0.013)  \\ 
  \hline
  X-coord & 0.770 ($\pm$ 0.013) &   0.104 & 0.0981\\ 
  Y-coord & 0.750 ($\pm$ 0.017) & 0.011 & 0.007\\
  Size & 0.812 ($\pm$ 0.009) &  0.315 & 0.47 \\
  Aspect ratio & \textbf{0.494} ($\pm$ 0.010) &0.916 & 0.989 \\
  RLF & 0.786 ($\pm$ 0.021) & - & -\\
  \hline
\end{tabular}
\end{subtable}

\begin{subtable}[t]{0.48\textwidth}
\centering
\caption{Hippocampus AD classification results}\label{hippo}
\begin{tabular}{  |l | c | c |c|} 
  \hline
  \multirow{2}{1em}{Feature}& \multirow{2}{6em}{AAC (ours) (mean $\pm$ std)} &\multicolumn{2}{c|}{CaCE score} \\
  \cline{3-4}
  & & $25\%/75\%$ & $5\%/95\%$\\
  %\hline
  %Original & 0.821 ($\pm$ 0.020)  \\
  %\hline
  %Reconstructed & 0.816 ($\pm$ 0.028) \\ 
  \hline
  Volume & \textbf{0.530} ($\pm$ 0.027)& 0.407 & 0.885 \\ 
  Aspect ratio & 0.798 ($\pm$ 0.029)& 0.076 & 0.221\\
  Avg. bright.& 0.806 ($\pm$ 0.026)& 0.094 & 0.296 \\
  RLF & 0.816 ($\pm$ 0.025)& - & - \\
  \hline
\end{tabular}
\end{subtable}

\begin{subtable}[t]{0.485\textwidth}
\centering

\caption{Cell nucleus type classification results (CaCE not applicable)}\label{cell}
\begin{tabular}{ | l | c |} 
  \hline
  Feature & AAC (ours) (mean $\pm$ std) \\ 
  %\hline
  %Original & 0.606 ($\pm$ 0.004)  \\
  %\hline
  %Reconstructed & 0.581 ($\pm$ 0.002) \\ 
  \hline
  Size & {\bf 0.449} ($\pm$ 0.013) \\ 
  Aspect ratio& 0.519 ($\pm$ 0.007) \\ 
  Saturation & {\bf 0.468} ($\pm$ 0.003) \\
  Hue &  {\bf 0.482} ($\pm$ 0.011) \\
  RLF & 0.524 ($\pm$ 0.010) \\
  \hline
\end{tabular}
\end{subtable}

%\vspace{-4mm}
\end{figure}

Next, we perform the feature dimension collapse and report the accuracy after collapse (AAC) along with CaCE scores on corresponding features in Table \ref{results}. The performance drop when collapsed along random linear feature (RLF) dimensions is used as a baseline comparison for our method. In the ellipse dataset, we can see that collapsing the aspect ratio changes the accuracy to around random chance. This is exactly what we would hope because this is the only feature that separates the two classes. Collapsing other features slightly affects the performance, but is similar to or less than RLF, from which we can conclude the model does not depend on them.
In the hippocampus experiment, the model depends heavily on volume, almost dropping to random chance when volume is removed.  The other features (aspect ratio, brightness) do not substantially influence the classifier. Note that despite the well-known fact that AD reduces hippocampal volume, because of the black-box nature of deep neural networks, we don't know if volume is a feature that the classifier has learned. However, our AAC measure confirms that volume is an essential feature for the classifier to learn. 
Finally, for the nucleus dataset, the results suggest size, saturation, and hue features are being used by the classifier to identify cell types (note here that random chance is 1/6 = 0.167). 

While CaCE scores generally align with our method regarding the identification of important features, the scale does not necessarily indicate how critical these features are. For instance, in the ellipse dataset, the aspect ratio is the sole distinguishing feature between the two classes; however, the size feature receives a notably high CaCE score.
This could be in part due to the conditional VAE's inability to constrain the feature values effectively as elaborated earlier. 
Moreover, applying CaCE to continuous features poses challenges, as varying feature values leads to unpredictable impacts. Also, note CaCE is only defined for binary classifiers, and thus we do not compare it to AAC on the multi-class cell nuclei dataset.%Additionally, when utilized in a multi-class classifier, CaCE can only provide a singular score for each class, lacking a straightforward method to combine into a single result.

Finally, we performed an ablation study for the hippocampus dataset to test the importance of using the VAE model to restrict the feature collapse to the data manifold. We repeated the hippocampus experiment, but with feature collapse directly in the data space, i.e., by integrating gradients of the features using Equation \eqref{eq:integrate} in the ambient data space. The resulting accuracy after collapsing volume is 0.772 ($\pm$ 0.043). And the results for aspect ratio and average brightness are 0.814 ($\pm$ 0.021) and 0.819 ($\pm$ 0.021), respectively. We can see that the performance drop in volume is much less drastic and also much more variable across test/train splits. This indicates that the model's behavior become more unpredictable when collapsing features in the ambient data space, perhaps due to the images being off of the data manifold.
%which is understandable given that the resulting images will possibly be outside the data distribution on which the classifier was trained.

In conclusion, our method effectively captures classifier feature dependencies, emphasizing the assessment of feature significance rather than mere relevance. It's important to note that our method necessitates a large sample size for VAE training and relies on access to feature gradients. We plan to address these challenges in future research.

%\section{Conclusion and Future Work}

\section{Compliance with Ethical Standards}
This research study was conducted retrospectively using
human subject data made available in open access by LaMontagne et al.~\cite{lamontagne2019oasis} and Graham et al.~\cite{graham2021lizard}.
Ethical approval was not required as confirmed by the
license attached with the open access data.

\section{Acknowledgements}
This work was partially supported by NSF Smart and Connected Health grant 2205417.

\bibliographystyle{IEEEbib}
\bibliography{refs}

\end{document}